\documentclass[letterpaper]{article} 
\usepackage{aaai25}  
\usepackage{times}  
\usepackage{helvet}  
\usepackage{courier}  
\usepackage[hyphens]{url}  
\usepackage{graphicx} 
\usepackage{natbib}  
\usepackage{caption} 
\usepackage{algorithm}
\usepackage{algorithmic}

\usepackage{multirow}
\usepackage{booktabs} 
\usepackage{amssymb}
\usepackage{relsize}
\usepackage{amsmath} 
\usepackage{amsfonts}

\usepackage{subfigure}

\newtheorem{definition}{Definition}[section]

\usepackage{newfloat}
\usepackage{listings}
\DeclareCaptionStyle{ruled}{labelfont=normalfont,labelsep=colon,strut=off} 
\floatstyle{ruled}
\newfloat{listing}{tb}{lst}{}
\floatname{listing}{Listing}
\title{
    \textsc{DepressionX}: Knowledge Infused Residual Attention for Explainable Depression Severity Assessment
}

\author{
    Yusif Ibrahimov \textsuperscript{\rm 1},
    Tarique Anwar \textsuperscript{\rm 1},
    Tommy Yuan \textsuperscript{\rm 1}
}
\affiliations{
    \textsuperscript{\rm 1}Department of Computer Science\\
    University of York\\
   York, United Kingdom\\
    \{yusif.ibrahimov,tarique.anwar,tommy.yuan\}@york.ac.uk
}

\usepackage{bibentry}

\begin{document}

\maketitle

\begin{abstract}
In today's interconnected society, social media platforms have become an important part of our lives, where individuals virtually express their thoughts, emotions, and moods. These expressions offer valuable insights into their mental health. This paper explores the use of platforms like Facebook, $\mathbb{X}$ (formerly Twitter), and Reddit for mental health assessments. We propose a domain knowledge-infused residual attention model called \textsc{DepressionX} for explainable depression severity detection. Existing deep learning models on this problem have shown considerable performance, but they often lack transparency in their decision-making processes. In healthcare, where decisions are critical, the need for explainability is crucial. In our model, we address the critical gap by focusing on the explainability of depression severity detection while aiming for a high performance accuracy. In addition to being explainable, our model consistently outperforms the state-of-the-art models by over 7\% in terms of $\text{F}_1$ score on balanced as well as imbalanced datasets. Our ultimate goal is to establish a foundation for trustworthy and comprehensible analysis of mental disorders via social media.
\end{abstract}

%

\section{Introduction}\label{sec:intro}

Depression affects over 280 million people globally, with severe outcomes, including approximately 700,000 suicides annually \cite{who2024depression}. While treatments exist, barriers such as stigma and access leave over 70\% untreated \cite{olfson2016treatment}. The COVID-19 pandemic has intensified this crisis, highlighting the urgent need for effective and accessible methods for early symptom identification \cite{shader2020covid19}. Social media platforms like Facebook, $\mathbb{X}$, and Reddit have become pivotal spaces where users express mental states, providing valuable data for automated mental health assessments \cite{anwar2022tracking}. Traditional methods, such as interview and questionnaire-based methods, capture depression patterns but are resource-intensive and may lack scalability \cite{park2012depressive,park2013perception}.

Recent work increasingly uses social media-based models to detect depression \cite{ibrahimov2024explainable}. \citeauthor{shen2017depression} 
 (\citeyear{shen2017depression}) utilised a depression-labelled $\mathbb{X}$ dataset for multimodal analysis, while \citeauthor{sampath2022data} 
  (\citeyear{sampath2022data}) examined Reddit for severity-based classifications with machine learning models. Advanced deep learning approaches have emerged, such as model developed by \citeauthor{naseem2022early} 
  (\citeyear{naseem2022early})  with attention-based severity detection, or DepressionNet \cite{zogan2021depressionnet}, which integrates text summarisation to enhance performance.

\noindent\textbf{\textit{Existing limitations.}} Two major gaps persist in the literature. First, most studies frame depression detection as a binary task \cite{shen2017depression,sadeque2018measuring,zogan2021depressionnet}, overlooking severity levels critical for clinical interventions \cite{ibrahimov2024explainable}. Initial efforts address this, with some using regression or classification for severity; however, these models often lack domain-specific depth \cite{naseem2022early}. Second, the opacity of deep learning models complicates their application in healthcare. Transparent, explainable models are vital for building trust with users and professionals alike \cite{ibrahimov2024explainable}.

\noindent\textbf{\textit{Our research.}} We introduce \textsc{DepressionX}, a novel model for explainable depression severity detection using social media data. It leverages four representations—\textit{word}, \textit{sentence}, \textit{post}, and \textit{knowledge graph}—to capture textual and domain-specific features. Domain knowledge from Wikipedia is incorporated via a graph-based approach to contextualize depression-specific content. Severity levels are classified through ordinal regression, enhancing the model's ability to recognize ordered severity classes. Additionally, we provide model explainability insights to highlight decision-making transparency. \textsc{DepressionX} is both explainable and effective at identifying severity levels, filling a key gap in current research. Our contributions include:

\begin{enumerate}
    \item We propose a robust deep learning model, \textsc{DepressionX}, for explainable depression severity prediction from social media posts. 
    \item Our knowledge graph representation within \textsc{DepressionX} effectively captures contextual relations from Wikipedia articles with the help of a domain-specific knowledge-base. 
    \item We conduct extensive experiments and analyse our results from various perspectives to demonstrate our efficacy.
    \item We also shed light on the underlying decision-making processes of our model to demonstrate its explainability.
\end{enumerate}

The remainder of this paper is structured as follows. Section \ref{sec:relatedwork} covers related work, followed by the problem statement in Section \ref{sec:problemstatement}. Section \ref{sec:proposedmodel} introduces \textsc{DepressionX}, and experimental results are presented in Section \ref{sec:experimentalresults}. Finally, Section \ref{sec:conclusion} concludes the paper.

\section{Related Work} \label{sec:relatedwork}

\noindent\textbf{\textit{Depression detection from social media}}. Social media platforms are valuable resources for detecting mental disorders such as depression \cite{zogan2021depressionnet,naseem2022early}, eating disorders \cite{abuhassan2023ednet,abuhassan2023classification} and suicide ideation \cite{naseem2022hybrid,sawhney2021towards}. eRisk shared task \cite{losada2018overview,losada2017erisk} by the Conference and Labs of the Evaluation Forum (CLEF)  focused on automatic depression detection in user posts. Other studies have analysed depression indicators through emotions \cite{park2012depressive,park2013perception}, linguistic style \cite{trotzek2018utilizing}, social interactions \cite{mihov2022mentalnet,pirayesh2021mentalspot}, and online behaviours \cite{de2013predicting,shen2017depression}. For instance, \citeauthor{park2012depressive} 
 (\citeyear{park2012depressive}) used interviews to link $\mathbb{X}$ users' language with depressive moods, while 
  \citeauthor{shen2017depression} 
 (\citeyear{shen2017depression}) developed a labelled dataset from $\mathbb{X}$ for multimodal analysis. Deep learning (DL) models, including CNNs 
  \cite{jacovi2018understanding}, LSTMs 
  \cite{hochreiter1997long}, and GRUs \cite{Chung2014-ss}, have become effective tools in this domain. \citeauthor{trotzek2018utilizing} 
  (\citeyear{trotzek2018utilizing}) explored CNNs with word embeddings and linguistic style, while 
  \citeauthor{zogan2021depressionnet} 
  (\citeyear{zogan2021depressionnet}) achieved high performance using automatic text summarisation. However, most approaches are limited to binary classification, ignoring severity levels. \citeauthor{naseem2022early} 
 (\citeyear{naseem2022early}) recently introduced a multi-class model with TextGCN and BiLSTM for severity detection. Graph neural networks (GNNs) have also been applied in mental disorder detection, capturing textual and social relationships \cite{naseem2022early,Liu2021-uk,pirayesh2021mentalspot}. \citeauthor{mihov2022mentalnet} 
  (\citeyear{mihov2022mentalnet}) proposed MentalNet, a GCN-based model utilizing heterogeneous graphs, while 
  \citeauthor{pirayesh2021mentalspot} 
  (\citeyear{pirayesh2021mentalspot}) introduced MentalSpot, based on friends' social posts and interactions.

\noindent\textbf{\textit{Explainability and depression detection}}. While DL models are effective, their complexity often limits transparency—a critical factor in healthcare. Explainable models like decision trees are self-explanatory, but DL models require post-hoc explainability techniques to clarify decisions \cite{Arya2019-yg,Jia2022-hp}. In depression detection, explainability is often underexplored, despite the disorder's complexity. Attention-based methods, including self-attention \cite{Amini2020-zk}, hierarchical attention \cite{han-etal-2022-hierarchical}, and multi-head attention models \cite{naseem2022hybrid}, have been used to address this need, providing insights into the model's reasoning.

\section{Problem statement} \label{sec:problemstatement}

This paper focuses on analysing a given set of social media posts, represented as $P = \{p_1, p_2, \ldots, p_{N}\}$, authored by corresponding users $U = \{u_1, u_2, \ldots, u_{N}\}$. The objective is to utilise the textual content of user posts to estimate their severity levels of depression in accordance with clinical depression standards, including Disorder Annotation (DDA) and Beck’s Depression Inventory (BDI) \cite{beck_ii}. The spectrum of depression severity is categorised into four distinct levels, denoted as possible depression severity classes, $C = \{minimal, mild, moderate, severe\}$. Each post $p_i$ authored by $u_i$ is assigned to one of the four severity classes in the ground truth, i.e., $\forall p_i, \exists label_i \in C$. 
The research problem is to automatically estimate the depression severity of $p_i$ by classifying it into one of the four classes $c_j \in C$, i.e., $f(p_i)\rightarrow c_j, \mbox{ where } p_i\in P \mbox{ authored by } u_i\in U$.

\section{Proposed model} \label{sec:proposedmodel}

\begin{figure*}[!ht]
\begin{center}
 \includegraphics[width=\textwidth]{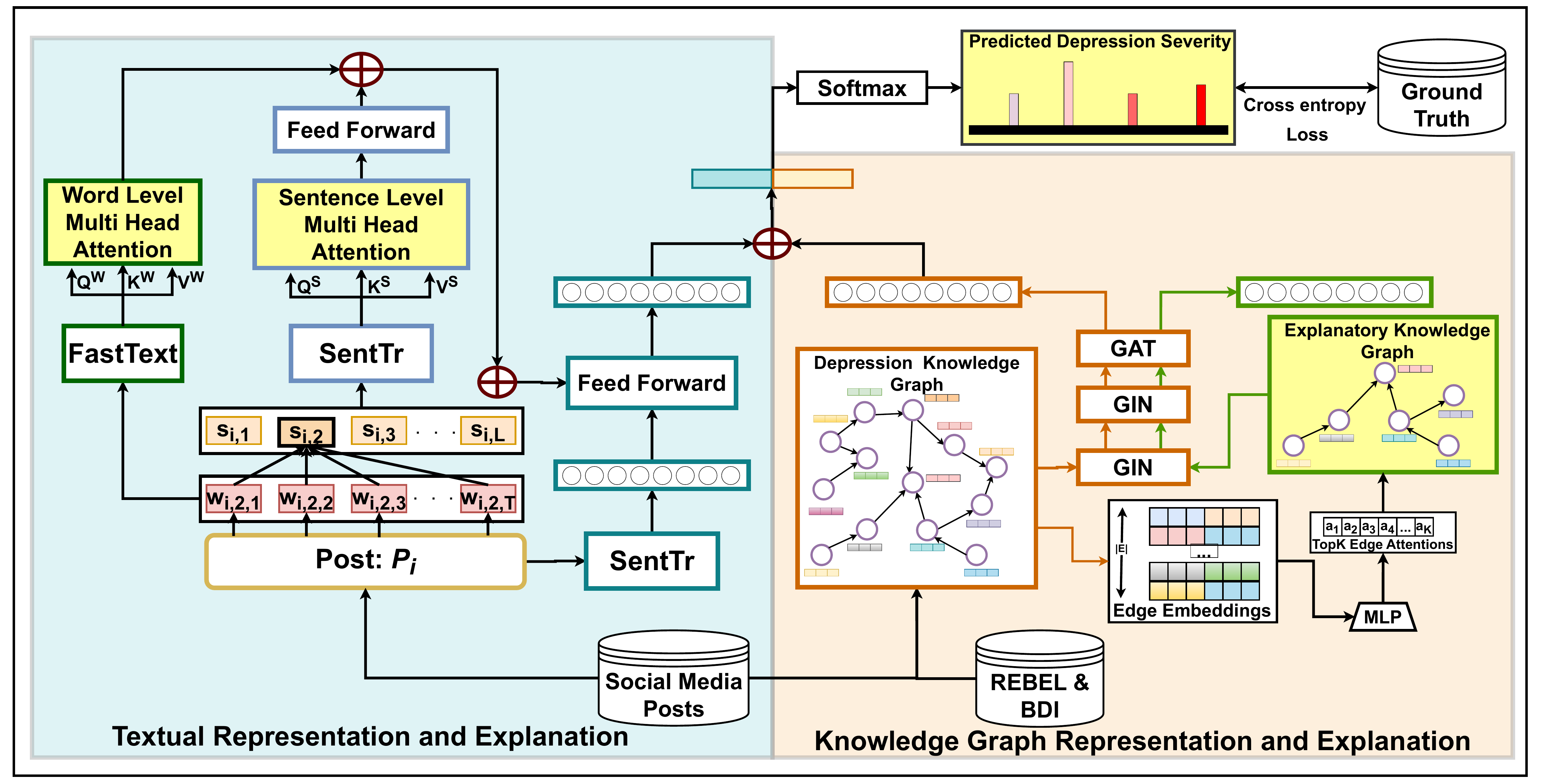}
  \caption{Proposed model \textsc{DepressionX}. It predicts depression severity of a post and generates explanation with word- and sentence-level attentions and a knowledge subgraph. The prediction and the explanatory components are highlighted in yellow.}
 \label{fig:depressionx}
\end{center}
\end{figure*}

The proposed model, \textsc{DepressionX}, learns a high-quality representation by integrating textual content with knowledge graph. The textual component extracts valuable information from user posts, while the knowledge graph component enriches this information with relevant domain knowledge. The architecture of the model is shown in Figure \ref{fig:depressionx} and discussed in detail in the remaining section below.

\subsection{Textual Representation}

\subsubsection{Multi-level Encoding}

Our model takes textual posts of social media platforms as input and encodes them at three different levels: \textit{word-level, sentence-level,} and \textit{post-level}. This is done to learn the representation at different granularities. For word-level encoding, we employ FastText \cite{bojanowski-etal-2017-enriching}, which is a computationally efficient pre-trained embedding model. It integrates neural networks with continuous bag of words (CBOW) and skip-gram processes. This method facilitates the representation of complex word structures through sub-words and n-grams, offering a novel approach to address out-of-vocabulary (OOV) words. Its ability to handle OOV words makes it particularly useful for our case study. This is because social media data often contains informal and distorted words, many of which are intentionally used by users to avoid violating social media rules. Specifically, we use the FastText 300-dimensional embeddings pre-trained on the English language. Thus the word-level encoding $\textbf{w}_{i,j,k}$ is represented as FastText$(
w_{i,j,k}) \in \mathbb{R}^{300}, \forall i\in \{1,N\}, \forall j \in \{1,L\}, \forall k \in \{1,T\}$, where $N$ is the total number of posts, $L$ is maximum number of sentences within a post and $T$ is the maximum number of words within a sentence.

For an effective sentence-level and post-level encoding, we adopt Sentence Transformer \cite{reimers2019sentence}. It efficiently generates a vector representation for each given sentence (or a document with multiple sentences) by applying a pooling operation on the output generated by transformer based language models such as RoBERTa \cite{liu2019roberta}. 
Sentence Transformers leverage Siamese \cite{bromley1993signature} and triplet networks to consider cosine similarities among documents, thereby updating weights to produce semantically meaningful contextual embeddings. Moreover,  they are computationally efficient and work significantly faster than other transformer based models such as RoBERTa. Thus, our sentence-level encoding $\textbf{s}_{i,j}$ within post $p_i$ is represented as $\text{SentenceTransformer}(s_{i,j}) \in \mathbb{R}^{768}, \forall i\in \{1,N\}, \forall j \in \{1,L\}$, and our post-level encoding $\textbf{p}_i$ is represented as $\text{SentenceTransformer}(p_i)\in \mathbb{R}^{768}, \forall i\in \{1,N\}$.

\subsubsection{Residual Multihead Attention}
To learn textual representation of a social media post $p_i$, we utilise a residual multi-head attention approach applied to multi-level encodings, defined as $\textbf{p}_i'  = \text{RMA}(p_i)$. In this approach, both sentence-level and post-level embeddings are enhanced by integrating them with word-level and combined word-sentence attentive representations. This method ensures that each post $p_i$ is encoded by comprehensively considering the significance of individual words and sentences, as well as the overarching context provided by the document embedding.

The attention mechanism is a component of neural architecture that dynamically emphasises crucial features of the input data. In our context, these features are represented by tokens, which may be sub-words, words, phrases, sentences, or documents. The core idea of attention is to compute a distribution of weights across the input sequence, assigning higher values to items deemed more relevant \cite{galassi2020attention}. To handle multiple modalities, capture complex relationships, and provide explainability through an analysis of attention weights highlighting the importance of words and sentences, we fine-tune a multihead attention model \cite{vaswani2017attention} for our task. This model is an enhancement over traditional single attention models. Attention scores are computed using Equation \ref{eqn:sa}, where $Q$ represents the \textit{query} (a trainable matrix for computing attention scores), $K$ stands for the \textit{key} (a trainable matrix on which attention weights are based), and $V$ denotes the \textit{value} (the matrix upon which computed attention is applied) \cite{galassi2020attention}. To capture diverse patterns and mitigate the risk of overfitting, we employ $k$ heads of attention operating in parallel called multihead attention (MHA) mechanism. Their outputs are concatenated to form the final representation vector as shown in Equation \ref{eqn:mha}, where $H_i = \text{Attention}(QW_i^Q,KW_i^K,VW_i^V)$.
%
%
\begin{equation}
    \text{Attention}(Q,K,V) = \text{softmax}  \left( \frac{QK^T}{\sqrt{d}}\right) \times V
     \label{eqn:sa}
\end{equation}
\begin{equation} \label{eqn:mha}
    \text{MHA}(Q,K,V) = \text{Concat}(H_1, H_2, \dots, H_k)\times W^0
\end{equation} 
Let us consider matrix $\mathbf{W}_{ij}$ generated by the word vectors $\{\textbf{w}_{i,j,k}\}_{k=1}^T$ of all words in a sentence $s_{i,j}$ from post $p_i$. 
We compute the corresponding attention values and further representation by applying MHA to the word vectors in $\mathbf{W}_{ij}$ using Equation \ref{eqn:mhasentence}, where $\mathbf{W'}_{ij}$ and $\Gamma_{\mathbf{W'}_{ij}}$ denote the overall word-level representation and word-level attention values, respectively.

\begin{equation} \label{eqn:mhasentence}
    \mathbf{W'}_{ij}, \Gamma_{\mathbf{W'}_{ij}} = \text{MHA}(Q_{\mathbf{W}_{ij}},K_{\mathbf{W}_{ij}},V_{\mathbf{W}_{ij}})
\end{equation} 
Thus, initially each post $p_i$ is represented as $\{\mathbf{W'}_{ij}\}_{j=1}^L$ using the word-level encoding. Subsequently, we compute the attention values and representations at the sentence level, employing the same logic for post $p_i$. The sentence vectors $\{\mathbf{s}_{i,j}\}_{j=1}^L$ are used to form matrix $\mathbf{S}_i$. The sentence representations and their attention values are obtained by applying multihead attention using Equation \ref{eqn:mhapost}.
\begin{equation} \label{eqn:mhapost}
    \mathbf{S'}_i, \Gamma_{\mathbf{S'}_i} = \text{MHA}(Q_{\mathbf{S'}_i},K_{\mathbf{S'}_i},V_{\mathbf{S'}_i})
\end{equation}
The overall sentence-level representation for post $p_i$ is obtained using Equation \ref{eqn:overallsentence}, where $\oplus$ and $\text{FFN}(\cdot)$ denote concatenation and  feed-forward network, respectively.
\begin{equation} \label{eqn:overallsentence}
    \mathcal{S}_ i = \{ \mathbf{W'}_j\}_{j=1}^L \oplus \text{FFN}(\mathbf{S}_i' )
\end{equation}
The final representation of post $p_i$ is obtained by Equation \ref{eqn:overallpost}.
\begin{equation} \label{eqn:overallpost}
    \textbf{p}_i' = \mathcal{S}_ i \oplus \text{FFN}(\textbf{p}_i) 
\end{equation}

  
\subsection{Knowledge Graph Representation}
\setcounter{section}{4} 
\subsubsection{Domain-specific knowledge graph construction}

\begin{definition}[Knowledge Graph]
A depression knowledge graph $\mathcal{G}(V,E,A,\mathcal{F})$ consists of $m$ nodes. Here, $V$ and $E$ represent the nodes $\{v_l\}_{l=1}^{m}$ (depression-related entities) and edges $\{e_{j,k}\}$ (indicating relations between nodes $v_j$ and $v_k$), respectively. Edges are represented as triplets \( e_{ht} = \langle v_h, r, v_t \rangle \), where \( v_h \) and \( v_t \) are the head and tail nodes, and \( r \) denotes the relationship type. $A \in \{0,1\}^{m \times m}$ is the adjacency matrix where $a_{j,k} = 1$ if $e_{j,k} \in E$. $\mathcal{F} \in \mathbb{R}^{m \times 768}$ represents node embeddings, generated by applying a sentence transformer on each node entity's summary, i.e., $\forall f_v \in \mathcal{F}$, $f_v = \text{SentenceTransformer}(s_v)$ where $s_v$ is the summary of node $v$.
\end{definition}
\begin{figure}[!ht]
\centering
\includegraphics[width=0.45\textwidth]{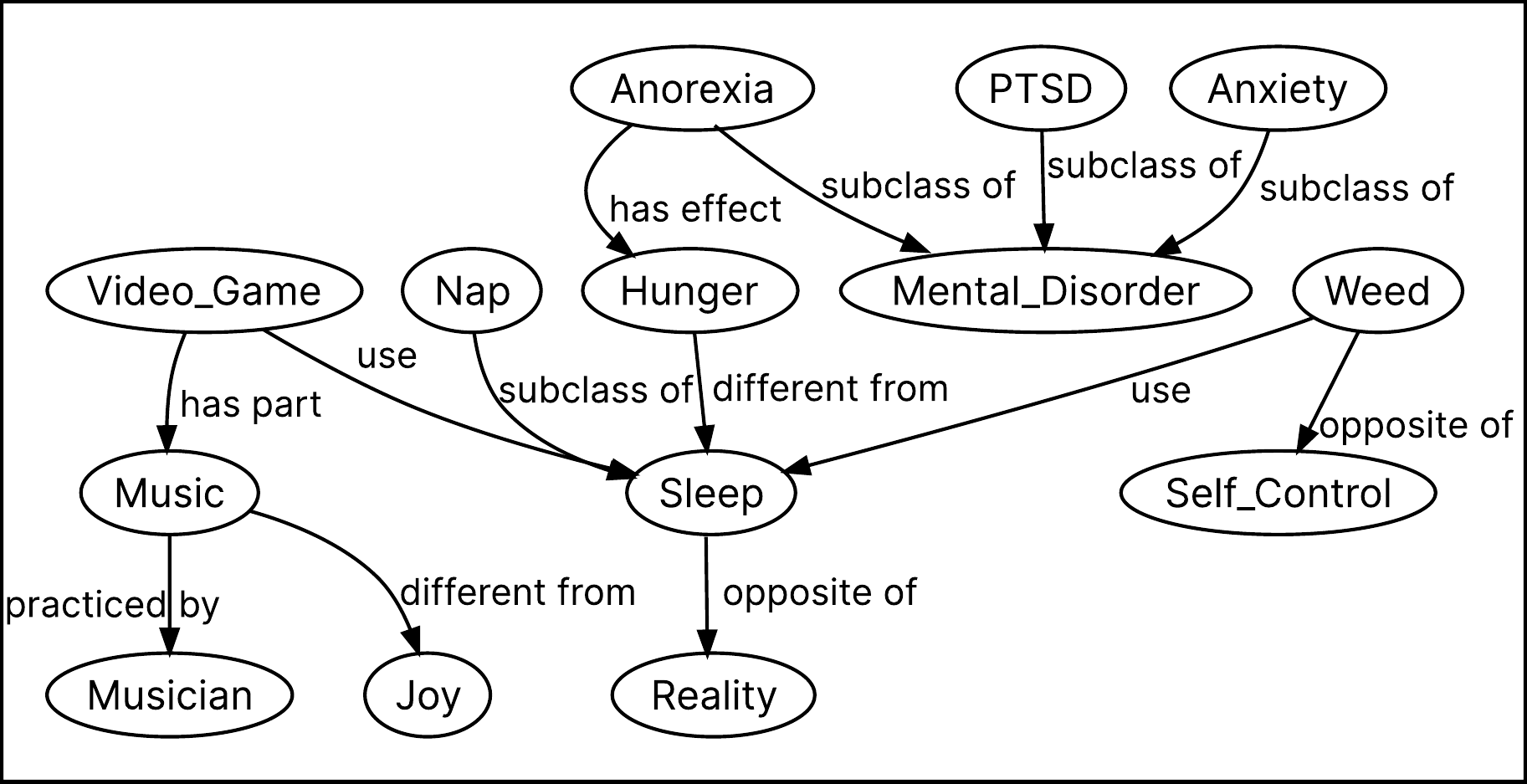}
\caption{Our domain-specific knowledge graph}
\label{fig:kg_dat}
\end{figure}
Knowledge graphs (KGs) are powerful in many retrieval applications by integrating external knowledge bases. Our method for depression severity detection involves constructing a domain-specific KG with relevant depression-related information and relationships. The KG construction starts with entity extraction using the Relation Extraction By End-to-end Language Generation (REBEL) model \cite{huguet-cabot-navigli-2021-rebel-relation}. REBEL reframes relation extraction as a sequence-to-sequence generation task, simplifying triplet extraction with BART, a transformer-based model \cite{Lewis2019-em}. Through REBEL, entities and relations are extracted as \textit{head}, \textit{tail}, and \textit{type} triplets from the text, leveraging REBEL's extensive pretraining on Wikipedia and Wikidata for high coverage and accuracy. We build our depression KG by feeding depression-related datasets \cite{sampath2022data} to REBEL, extracting entities, relationships, hyperlinks, and summaries from Wikipedia data. Afterward, entities (heads and tails) are refined to focus on depression by comparing their cosine similarity to depression symptoms based on the BDI-II \cite{beck_ii}, retaining entities with similarity scores above 0.5. These refined triplets structure the KG, capturing key relationships and contextual data on depression severity. Figure \ref{fig:kg_dat} shows an excerpt of our depression KG, depicting its structured representation of relevant information and relationships.

\subsubsection{Knowledge graph representation learning}

To capture structural properties in our depression knowledge graph (KG) and gain insights into depression severity, we use two sequential layers of graph isomorphism network (GIN) \cite{xu2018powerful} followed by a layer of graph attention network (GAT) \cite{velivckovic2017graph}. The GIN layers aggregate information from up to 2-hop neighbours, while the GAT layer focuses on the third-hop neighbours with attention, benefiting from GIN's strong representation capability and GAT's embedded attention mechanism.

GIN is a graph neural network designed to capture complex structural patterns, distinguishing graphs even with similar features. We process our KG $\mathcal{G}(V,E,A,\mathcal{F})$ with two GIN layers, as shown in Equation \ref{eqn:gin}, learning the embedding $h_v^{(k)}$ for each node $v \in V$ at layer $k$. Here, $h_v^{(0)} = f_v$, $\epsilon$ denotes the importance of the target node relative to its neighbours, and $\mathcal{N}(v)$ represents the neighbours of $v$.
\begin{equation}
    h_v^{(k)} = \text{MLP}\left((1 + \epsilon^{(k)}) \cdot h_v^{(k-1)} + \sum_{u \in \mathcal{N}(v)}h_u^{k-1}\right)
     \label{eqn:gin}
\end{equation}

Following GIN, we apply a multihead GAT model 
  \cite{velivckovic2017graph} to mitigate fixed edge weights in GIN, incorporating an attention mechanism that enables selective aggregation from neighbours based on their importance. GAT assigns each neighbouring node an attention coefficient $\alpha_{ij}$, computed using Equation \ref{eqn:gatattention}, where $||$ denotes concatenation, $h_i$ and $h_j$ are the hidden representations of nodes $i$ and $j$, and $\textbf{a}$ and $\Theta$ are trainable parameters.
\begin{equation} \label{eqn:gatattention}
        \alpha_{ij} = \text{softmax} \left(\text{LeakyReLU}\left({\textbf{a}}^T\left[\Theta h_i^{(2)} \Vert \Theta h_j^{(2)}\right]\right)\right) 
\end{equation}

neighbouring nodes $j \in \mathcal{N}_i$ contribute to node $i$’s updated feature $h_i'$ based on $\alpha_{ij}$ and $\Theta$ with a non-linear activation $\sigma$. GAT employs multihead attention (Equation \ref{eqn:gatmultihead}) for stability, where $\alpha_{ij}^k$ is the attention coefficient from the $k$-th head, and $\Theta^k$ is the associated weight matrix, allowing the model to capture various aspects of relationships.
    \begin{equation} \label{eqn:gatmultihead}
    h_i^{(3)} = \sigma \left( \frac{1}{K} \sum_{k=1}^K \sum_{j \in \mathcal{N}_i} \alpha_{ij}^k \Theta^kh_j^{(2)}\right)
\end{equation}
Finally, we apply max pooling over nodes to obtain the learned KG representation:
\begin{equation} \label{eqn:maxpool}
    \textbf{g} = h(G) = \text{MaxPool}(\{h_i^{(3)}\}_{i=1}^{m}) 
\end{equation}

\subsection{Knowledge Infusion and Output}
To leverage the complementary information from both social media posts and the KG, we concatenate their learned representations for the final depression severity prediction using  Equations \ref{eqn:fusion} and \ref{eqn:op}.
Here $f(p_i)$ is the final depression severity predicted for post $p_i$ and  $\text{FFN}(\cdot)$ indicates a feed-forward network.

\begin{eqnarray}
    \label{eqn:fusion} \textbf{z}_i = \textbf{p}_i' \oplus \textbf{g} \\
    \label{eqn:op} f(p_i) = \text{argmax } \text{softmax}\left(\text{FFN}(\textbf{z}_i)\right)
\end{eqnarray} 

Considering the ordinal relation among depression severity levels, we adopt ordinal regression inspired by 
 \citeauthor{sawhney2021towards} (\citeyear{sawhney2021towards}). Let $\mathcal{Y} = \{ minimum=0, mild=1, moderate=2, severe=3 \} = \{ r_i \}_{i=0}^4$ be the labels of depression severity in an ascending order. For soft label of the severity $r_p \in \mathcal{Y}$ of a post $p$, we calculate probability distribution $\textbf{y}^p = [y_0^p,y_1^p,y_2^p,y_3^p]$ using Equation \ref{eqn:softlabel}, where  $\phi(r_p,r_i)$ is a cost function that computes the distance between the actual severity level $r_p \in \mathcal{Y}$ and all the severity levels $r_i \in \mathcal{Y}$ in Equation \ref{eqn:cost}. Here, $\beta$  is a parameter that controls the magnitude of penalty for incorrectly predicting a severity. As the difference between $r_p$ and $r_i$ increases, there is a decrease in the probability $y_i^p$ of the associated prediction $r_i$.
\begin{equation} \label{eqn:softlabel}
    y_i^p = \frac{\exp^{- \phi(r_p,r_i)}}{\sum_{r_k\in\mathcal{Y}}\exp^{- \phi(r_p,r_k)}}
\end{equation}
\begin{equation}
    \phi(r_p,r_i) = \beta|r_t - r_i|
    \label{eqn:cost}
\end{equation}
\subsection{Explainability}\label{sec:exp}

Neural networks encode knowledge in numerical formats that inherently lack explainability. This opacity raises ethical and practical challenges, especially in healthcare, where understanding model decisions is crucial for trust and accountability. Addressing this, we propose methods to elucidate the decision-making process of \textsc{DepressionX}, leveraging the attention mechanism for improved explainability \cite{galassi2020attention}.

Our approach uses attention values to explain both text and knowledge graph (KG) representations. Specifically, we analyse the influence of words, sentences, and KG components on the model's predictions. For text, \textsc{DepressionX}'s multi-head attention mechanism generates word- and sentence-level attention scores, quantifying their significance. However, explaining KG representations requires a more nuanced strategy due to their structural complexity.

To tackle this, inspired by \citeauthor{luo2020parameterized} 
 (\citeyear{luo2020parameterized}), we decompose the original KG, $G_O$, into an explanatory subgraph, $G_S$, and an irrelevant subgraph, $\Delta G$, such that $G_O = G_S \cup \Delta G$. Our goal is to maximize the mutual information between $G_O$ and $G_S$, ensuring $G_S$ retains critical information. This is formulated as an optimisation problem, minimising the loss between representations of $G_O$ and $G_S$:
\begin{equation}
\min_{G_S} \text{Loss}(f(G_O), f(G_S))
\end{equation}
where the loss function, defined as Smooth L1, balances mean squared and mean absolute errors:

Edge attention is computed using a multi-layer perceptron (MLP) trained on edge embeddings. Edge embeddings, $e_{ij}'$, are derived by concatenating feature vectors of adjacent nodes:
\begin{equation}
\forall e_{ij} \in \textbf{E}, \quad e_{ij}' = [\textbf{f}_{v_i}, \textbf{f}_{v_j}] \in \textbf{E'}
\label{eq:edge_emb}
\end{equation}
where $\textbf{E'} \in \mathbb{R}^{|\textbf{E}|\times 2 \times d}$. The MLP maps these embeddings to attention scores:
\begin{eqnarray}
\text{MLP}: \mathbb{R}^{|\textbf{E}|\times 2 \times d} \rightarrow \mathbb{R}^{|\textbf{E}|} \\
\text{MLP}: \textbf{E'} \mapsto \textbf{W}_y \times \textbf{E'} + \textbf{b}_y
\label{eq:mlp}
\end{eqnarray}
where $\textbf{W}_y$ and $\textbf{b}_y$ are trainable parameters. We select the top-K edges with the highest attention scores to construct $G_S$. This method yields a compact, explainable subgraph that highlights key components influencing \textsc{DepressionX}'s decisions.

\section{Experimental Results} \label{sec:experimentalresults}

\subsection{Datasets}
We evaluate the performance of \textsc{DepressionX} using two benchmark datasets (\textbf{D1} and \textbf{D2}) from Reddit, developed with field experts and adhering to The Diagnostic and Statistical Manual of Mental Disorders (DSM) classification criteria \cite{apa2013diagnostic}. Our analysis is focused on Reddit data, as current depression severity datasets are exclusive to this platform. However, our embedding methods are adaptable to data from any data sources. To assess the robustness of our model, we selected datasets differing in class distribution: \textbf{D1} is slightly imbalanced, while \textbf{D2} is more balanced. This allows evaluation across both scenarios. \textbf{D1} \cite{naseem2022early}: This dataset includes posts categorized into \textit{non-depressed/minimum, mild, moderate,} and \textit{severe} classes, comprising 2587, 290, 394, and 282 posts, respectively. It is an augmented version of the Dreaddit dataset \cite{turcan-mckeown-2019-dreaddit}, refined using DSM-V \cite{apa2013diagnostic} standards. \textbf{D2} \cite{sampath2022data}: Comprising 1985, 1000, and 902 posts in the \textit{non-depressed, moderate,} and \textit{severe} categories, respectively, this dataset is sourced from subreddits such as ``r/MentalHealth,'' ``r/depression,'' and others, using the Reddit API.

\subsection{Experimental Settings}

We determined the optimal hyperparameters based on the highest weighted $\text{F}_1$ score by employing the Optuna framework (https://optuna.org/) with the Tree-structured Parzen Estimator sampler (TPE sampler). The values are: learning rate of 0.000278, 100 epochs, 8 heads, dropout rate of 0.4, contrastive loss temperature of 0.5, batch size of 4, severity scale of 3.0, hidden size of 128, maximum length of 64, and cosine similarity score of 0.5. The Top-K values for the explanatory Knowledge Graphs (KGs) depend on the preferences of the individual performing the analysis.

We fine-tuned the language models using Huggingface's Transformer library. The implementation is done in PyTorch 2.1, utilising the Adam optimiser.\footnote{The code and data are publicly available at the following link: \url{https://github.com/ioseff-i/DepressionX}.} All hyperparameter optimisation was performed exclusively on the training and validation splits of the datasets, with no access to the test data. Our experiments are conducted on the Viking HPC Cluster provided by the University of York, providing access to 12,864 CPU cores, 512GB of RAM, and NVidia A40 GPUs. Model evaluation is performed using weighted precision, recall and $\text{F}_1$ score. To assess the reliability and stability of model performance, each model is repeatedly run 10 times, and the median of the evaluation measures along with standard deviations are reported.



\subsection{Compared Models}
We evaluate \textsc{DepressionX} against six generic baselines (LSTM, GRU, BiLSTM, BiGRU, CNN, BERT) and five state-of-the-art (SOTA) depression detection models: MentalBERT \cite{Ji2021-hu}, DepressionNet 
  \cite{zogan2021depressionnet}, model developed by 
 \citeauthor{naseem2022early} (\citeyear{naseem2022early}), by \citeauthor{ghosh2021depression} 
  (\citeyear{ghosh2021depression}), and MuLHiTA \cite{xia2022mulhita}. All models are adapted for depression severity prediction by modifying their output layers. LSTM, GRU, BiLSTM, and BiGRU are recurrent networks tailored for sequential data. CNNs extract salient local features through convolution. BERT and MentalBERT, both transformer-based models, are influential in NLP, with MentalBERT trained specifically on mental health data. DepressionNet performs binary depression classification using text summarisation, BERT embeddings, BiGRU, attention, and auxiliary social media features.  \citeauthor{naseem2022early} (\citeyear{naseem2022early}) utilise TextGCN, BiLSTM, and attention for ordinal regression of depression severities. \citeauthor{ghosh2021depression} 
  (\citeyear{ghosh2021depression}) predict depression intensity using LSTM enhanced with features like emotion, topic, behaviour, and user traits.  MuLHiTA \cite{xia2022mulhita}, originally designed for EEG data, applies hierarchical temporal attention and BiLSTM on segmented data slices to assess mental stress. We adapt MuLHiTA for text by treating EEG slices as sentence segments from social media posts.

\subsection{Performance Comparison}
\begin{table*}[!h]\centering
\caption{Overall performance results for D1 and D2 over 10 runs (median of the measure $\pm$ standard deviation)}
\label{tab:combined_results}
\scalebox{0.75}{
\begin{tabular}{p{5.0cm}|ccc cccc}
\toprule
\multirow{2}{*}{\textbf{Model}} & \multicolumn{3}{c}{\textbf{D1}} & & \multicolumn{3}{c}{\textbf{D2}} \\
\cmidrule{2-4} \cmidrule{6-8}
& \textbf{Precision} & \textbf{Recall} & \textbf{$\text{F}_1$ Score} & $\quad$& \textbf{Precision} & \textbf{Recall} & \textbf{$\text{F}_1$ Score} \\
\midrule
LSTM & 0.622 $\pm$ 0.085 & 0.724 $\pm$ 0.022 & 0.649 $\pm$ 0.044 & & 0.779 $\pm$ 0.041 & 0.774 $\pm$ 0.041 & 0.776 $\pm$ 0.048 \\
GRU & 0.555 $\pm$ 0.022 & 0.685 $\pm$ 0.034 & 0.609 $\pm$ 0.019 & & 0.798 $\pm$ 0.028 & 0.789 $\pm$ 0.029 & 0.784 $\pm$ 0.031 \\
BiLSTM & 0.696 $\pm$ 0.065 & 0.727 $\pm$ 0.023 & 0.677 $\pm$ 0.039 & & 0.705 $\pm$ 0.037 & 0.711 $\pm$ 0.037 & 0.700 $\pm$ 0.039 \\
BiGRU & 0.688 $\pm$ 0.068 & 0.703 $\pm$ 0.032 & 0.661 $\pm$ 0.042 & & 0.784 $\pm$ 0.035 & 0.767 $\pm$ 0.048 & 0.762 $\pm$ 0.044 \\
CNN & 0.746 $\pm$ 0.068 & 0.762 $\pm$ 0.028 & 0.726 $\pm$ 0.049 & & 0.821 $\pm$ 0.038 & 0.818 $\pm$ 0.038 & 0.817 $\pm$ 0.039 \\
BERT \cite{devlin2018bert}& 0.530 $\pm$ 0.001 & 0.695 $\pm$ 0.001 & 0.610 $\pm$ 0.001 & & 0.590 $\pm$ 0.001 & 0.846 $\pm$ 0.001 & 0.675 $\pm$ 0.001 \\
MentalBERT \cite{Ji2021-hu} & 0.523 $\pm$ 0.001 & 0.698 $\pm$ 0.001 & 0.611 $\pm$ 0.001 & & 0.561 $\pm$ 0.001 & 0.811 $\pm$ 0.001 & 0.645 $\pm$ 0.001 \\
DepressionNet \cite{zogan2021depressionnet}& 0.686 $\pm$ 0.017 & 0.686 $\pm$ 0.017 & 0.678 $\pm$ 0.017 & & 0.817 $\pm$ 0.022 & 0.814 $\pm$ 0.016 & 0.808 $\pm$ 0.018 \\
\citeauthor{naseem2022early}(\citeyear{naseem2022early}) & 0.520 $\pm$ 0.033 & 0.714 $\pm$ 0.020 & 0.595 $\pm$ 0.057 & & 0.255 $\pm$ 0.021 & 0.505 $\pm$ 0.021 & 0.339 $\pm$ 0.023 \\
\citeauthor{ghosh2021depression} (\citeyear{ghosh2021depression}) & 0.515 $\pm$ 0.025 & 0.717 $\pm$ 0.017 & 0.599 $\pm$ 0.023 & & 0.565 $\pm$ 0.031 & 0.724 $\pm$ 0.046 & 0.625 $\pm$ 0.044 \\
MuLHiTA \cite{xia2022mulhita} & 0.753 $\pm$ 0.034 & 0.670 $\pm$ 0.012 & 0.707 $\pm$ 0.021 & & 0.839 $\pm$ 0.009 & 0.834 $\pm$ 0.009 & 0.835 $\pm$ 0.009 \\
\textbf{\textsc{DepressionX}} & \textbf{0.974 $\pm$ 0.048} & \textbf{0.718 $\pm$ 0.003} & \textbf{0.825 $\pm$ 0.020} & & \textbf{0.911 $\pm$ 0.006} & \textbf{0.909 $\pm$ 0.006} & \textbf{0.909 $\pm$ 0.006} \\
\bottomrule
\end{tabular}}
\end{table*}

Table \ref{tab:combined_results} shows the detailed results of all the compared models on \textbf{D1} and \textbf{D2}. Our model, \textsc{DepressionX}, consistently outperforms all competitors on both datasets. It achieves an 82.5\% $\text{F}_1$ score, 71.8\% recall, and 97.4\% precision score on \textbf{D1}, and a 90.9\% $\text{F}_1$ score, 90.9\% recall, and 91.1\% precision score on \textbf{D2}.
The RNN-based models, including LSTM, BiLSTM, GRU, and BiGRU, show notable results due to their sequential learning capabilities. However, they are surpassed by CNN, which captures advanced patterns with its convolution and pooling mechanisms. RNNs are prone to vanishing gradient issues. DepressionNet, a combination of BiGRU, CNNs, and an attention mechanism with contextual embedding, shows great performance. Nonetheless, \textsc{DepressionX} outperforms DepressionNet by incorporating residual multi-head attention and domain knowledge adaptation. Surprisingly, advanced models such as MentalBERT, BERT, and the method of 
 \citeauthor{naseem2022early} (\citeyear{naseem2022early}) perform worse than RNN-based and CNN models. The likely reason behind the weak performance of method proposed by 
  \citeauthor{naseem2022early} (\citeyear{naseem2022early}) is the lack of advanced text embedding techniques. We also observe a performance gap between the two datasets, primarily due to differences in label distribution. MuLHiTA \cite{xia2022mulhita} produces impressive results, surpassing most of the competing models, mainly because it comprehensively examines segments (sentences) from multiple perspectives using BiLSTM and temporal attention. Nevertheless, our model, \textsc{DepressionX}, exceeds the performance of all other models by more than 7\% in terms of weighted $\text{F}_1$ score on both datasets. This success is attributed to its consideration of multiple factors such as word-level, sentence-level, and post-level contextual embeddings, alongside domain-specific knowledge.

\subsection{Ablation Study}
We performed an ablation study to understand the impact of each component in \textsc{DepressionX}: word-level (Block 1), sentence-level (Block 2), and post-level embeddings (Block 3). Table \ref{tab:combined_ablation} summarises the results across 10 runs, showing the median and standard deviation of key metrics. 
Using only word-level embeddings (Block 1) provides strong performance on D1 due to the dataset's concise text but is less effective on the more balanced D2. Incorporating sentence-level (Block 2) and post-level embeddings (Block 3) significantly improves results, especially for D2, by capturing richer context. When combined, Blocks 1 and 2 offer marked gains, and adding Block 3 yields further improvements, demonstrating the advantage of multi-level representation. 
\begin{table*}[!h]\centering
\caption{Ablation study results over 10 runs (median $\pm$ standard deviation) for datasets D1 and D2.}\label{tab:combined_ablation}
\scalebox{0.9}{
\begin{tabular}{l|cc ccc}\toprule
\multirow{2}{*}{\textbf{Model}}  & \multicolumn{2}{c}{\textbf{D1}} && \multicolumn{2}{c}{\textbf{D2}} \\
\cmidrule{2-3} \cmidrule{5-6}
& \textbf{Weighted $\text{F}_1$} & \textbf{Accuracy} &$\quad$& \textbf{Weighted $\text{F}_1$} & \textbf{Accuracy} \\
\midrule
Word-level Embedding (Block 1) & 0.840 $\pm$ 0.007 & 0.726 $\pm$ 0.001 && 0.666 $\pm$ 0.010 & 0.518 $\pm$ 0.021 \\
Sentence-level Embedding (Block 2) $\qquad$ & 0.638 $\pm$ 0.146 & 0.643 $\pm$ 0.109 && 0.808 $\pm$ 0.024 & 0.808 $\pm$ 0.023 \\
Post-level Embedding (Block 3) & 0.667 $\pm$ 0.061 & 0.656 $\pm$ 0.039 && 0.770 $\pm$ 0.004 & 0.767 $\pm$ 0.004 \\
Block 1 + Block 2 & 0.798 $\pm$ 0.010 & 0.701 $\pm$ 0.078 && 0.873 $\pm$ 0.005 & 0.873 $\pm$ 0.005 \\
Block 1 + Block 2 + Block 3 & 0.835 $\pm$ 0.001 & 0.717 $\pm$ 0.001 && 0.885 $\pm$ 0.005 & 0.885 $\pm$ 0.004 \\
\textsc{DepressionX} (full) & 0.825 $\pm$ 0.020 & 0.718 $\pm$ 0.003 && 0.909 $\pm$ 0.006 & 0.909 $\pm$ 0.006 \\
\bottomrule
\end{tabular}
}
\end{table*}

The full \textsc{DepressionX} model, which integrates domain knowledge through a knowledge graph (KG), shows the best overall performance, with notable gains on D2. However, on D1, the additional complexity sometimes causes performance to dip slightly, likely because D1’s shorter posts do not benefit as much from higher-level contextual features, and embeddings may introduce redundancy or interference. Sentence-level embeddings outperform post-level ones by capturing detailed context more effectively, but a combined approach balances granularity and comprehensiveness.
\begin{figure}[!h]
\vspace{-5pt}
\centering
  \includegraphics[width=0.45\textwidth]{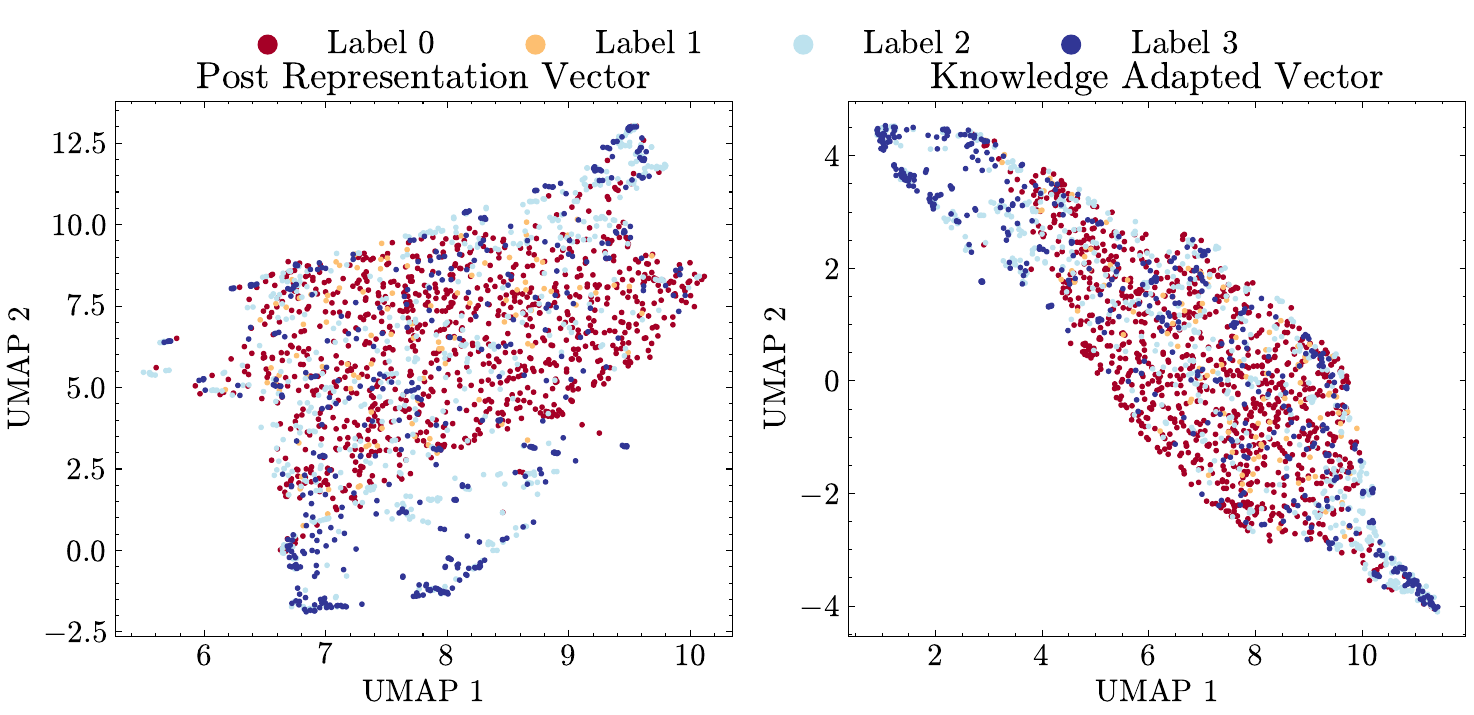}
  \vspace{-5pt}
  \caption{Two-dimensional plots of post representations (generated by UMAP) before and after knowledge infusion}
\label{fig:umap}
\vspace{-5pt}
\end{figure}
To evaluate the impact of knowledge infusion, we employ uniform manifold approximation and projection (UMAP) for dimensionality reduction of our learned representation. We plot the representations after reducing them into a 2-dimensional space against their severities both before and after the infusion of knowledge. Figure \ref{fig:umap} shows the effect of KG in \textbf{D1}. After infusion domain knowledge, we can clearly see a significant enhancement in the distinction between various classes. This suggests that the domain knowledge module assists in enhancing the discriminative nature of the embeddings, which is essential for recognizing patterns associated with the severity of depression. The map exhibits distinct boundaries and well-defined clusters, while there is still some dispersion within the same category. This visual evidence corroborates the assertion that the domain knowledge module has enhanced the depiction of the data. The intensity of depression is fundamentally a continuous spectrum rather than discrete nature. The variation within the same category can be ascribed to the subtle distinctions in the levels of severity that UMAP detects. This attribute is, in fact, a strength, as it accurately mirrors the intricate nature of the severity of depression in real life.

\subsection{Explainability Analysis} \label{sec:explainabilityanalysis}
\begin{figure}[!h]
\centering
  \includegraphics[width=0.45\textwidth]{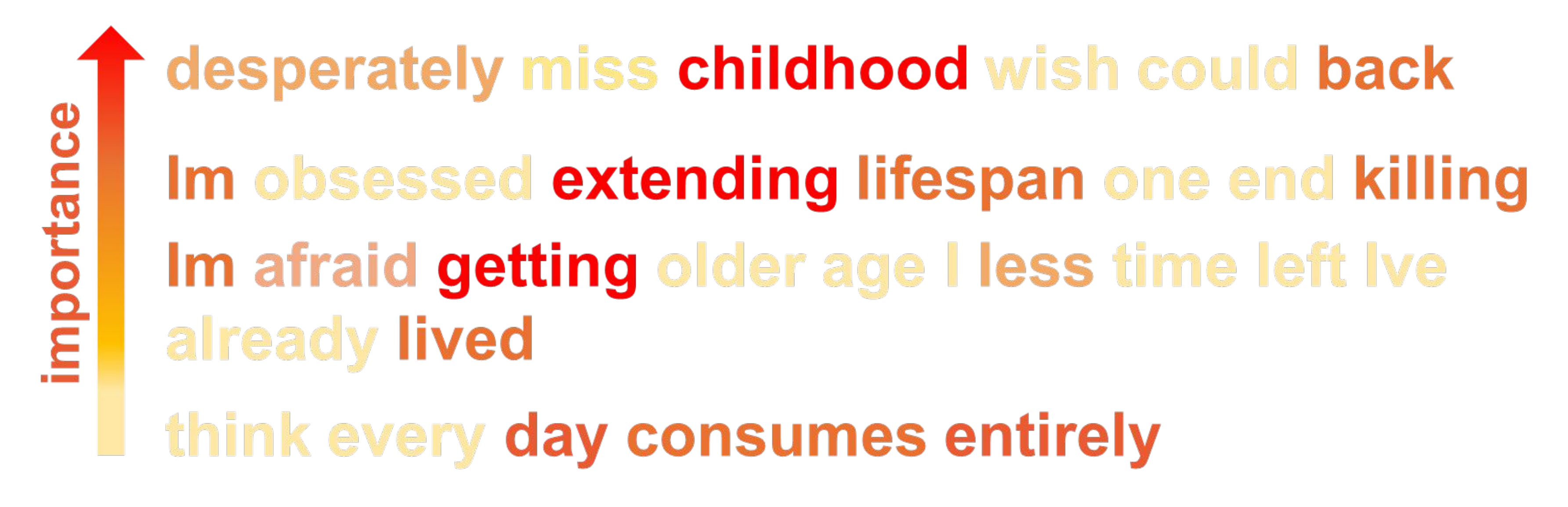}
  \caption{Explainability with sentence- and word-level importance}
\label{fig:text_exp}
\end{figure}

To explain the \textsc{DepressionX} model and its outcome, we conduct an analysis encompassing two distinct aspects: text and the KG. Figure \ref{fig:text_exp} demonstrates the explainability of our methodology with the textual content of individual posts. In our approach, we leverage word-level and sentence-level attention mechanisms to identify the salient features. By ranking sentences and words based on their attention values, we highlight the most pertinent components that influence the decision-making process.

\begin{figure*}[!ht]
\centering
  \includegraphics[width=0.9\textwidth]{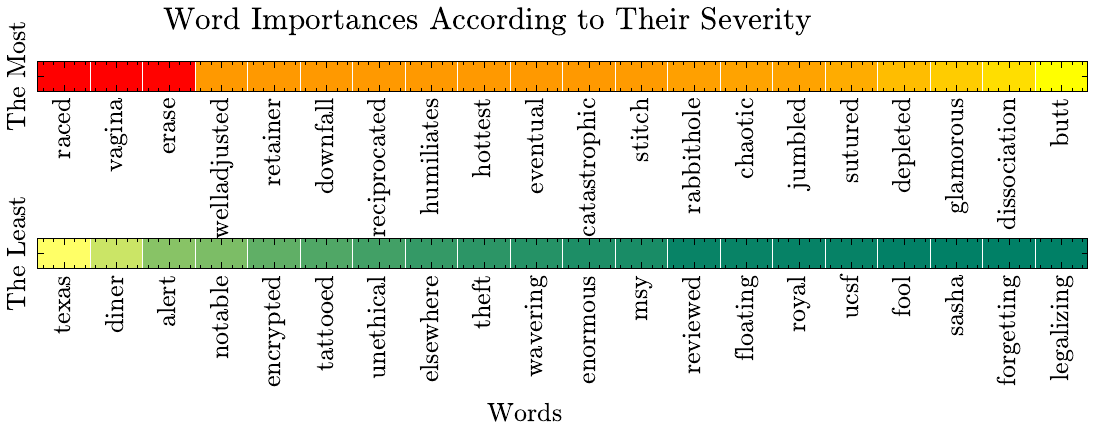}
  \caption{Attention heatmap of the most and the least important words}
\label{fig:word_att}
\end{figure*}

Figure \ref{fig:word_att} presents the heatmap of attention values for the most and least important words in the datasets. As observed in the figure, the most important words for the decision include abusive words, words with sexual references, slang, swear words, compulsive words, and words that evoke problems. Conversely, normal words with little emotional impact are deemed less important.

\begin{figure}[!ht]
\begin{center}
  \includegraphics[width=0.45\textwidth]{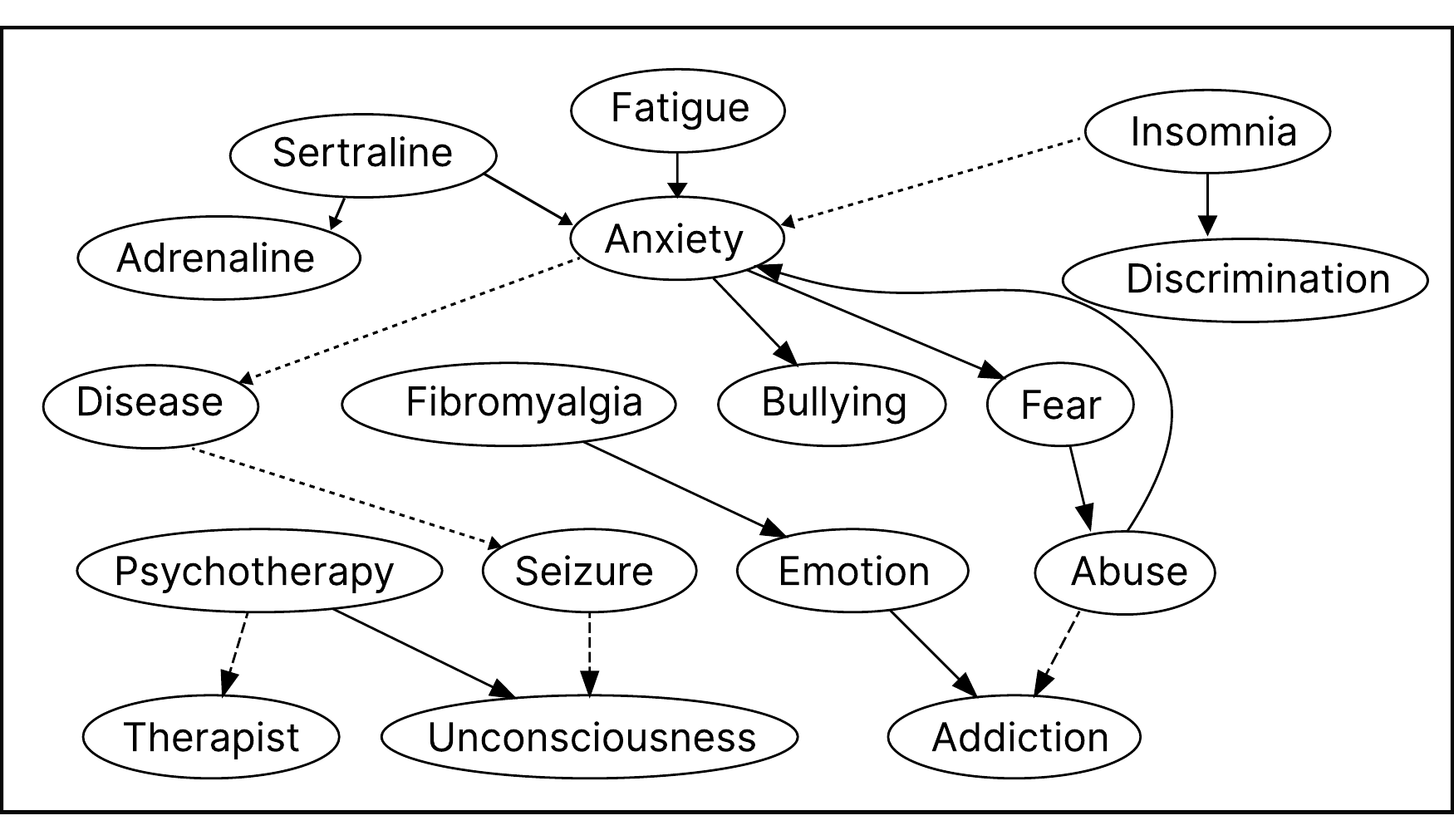}
  \caption{Explanatory knowledge subgraph}
\label{fig:sub_g}
\end{center}
\end{figure}

For explaining the KG, we employ edge importance measures, as discussed in Section \ref{sec:exp}. The original KG comprises over 100 nodes and edges, making direct analysis challenging. Therefore, we extract a minimal subgraph that maximises explainability. Figure \ref{fig:sub_g} illustrates such a subgraph, which maximises the mutual information with the original KG. In the figure, dashed arrows denote less significant edges, while solid-lined arrows signify the most crucial relationships within the graph. Examining the corresponding subgraph for a predicted outcome provides insights into the factors that led the model to reach its decision.

\subsection{Parametric Analysis}
\label{sec:parametricyanalysis}

\begin{figure}[!ht]
 \centering
 \subfigure[Left: learning rate, Right: number of heads in multihead attention (H) and number of heads in GAT (GH)]  
 {\includegraphics[width=0.45\textwidth]{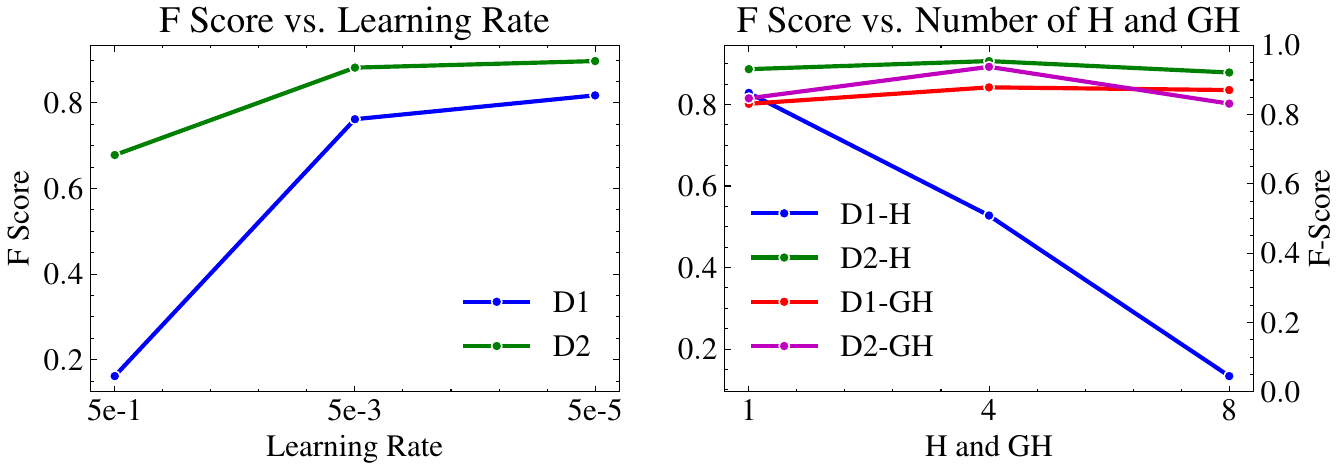} \label{fig:par_an}}
 \subfigure[Left: temperature value for contrastive loss, Right: maximum number of words] 
 {\includegraphics[width=0.45\textwidth]{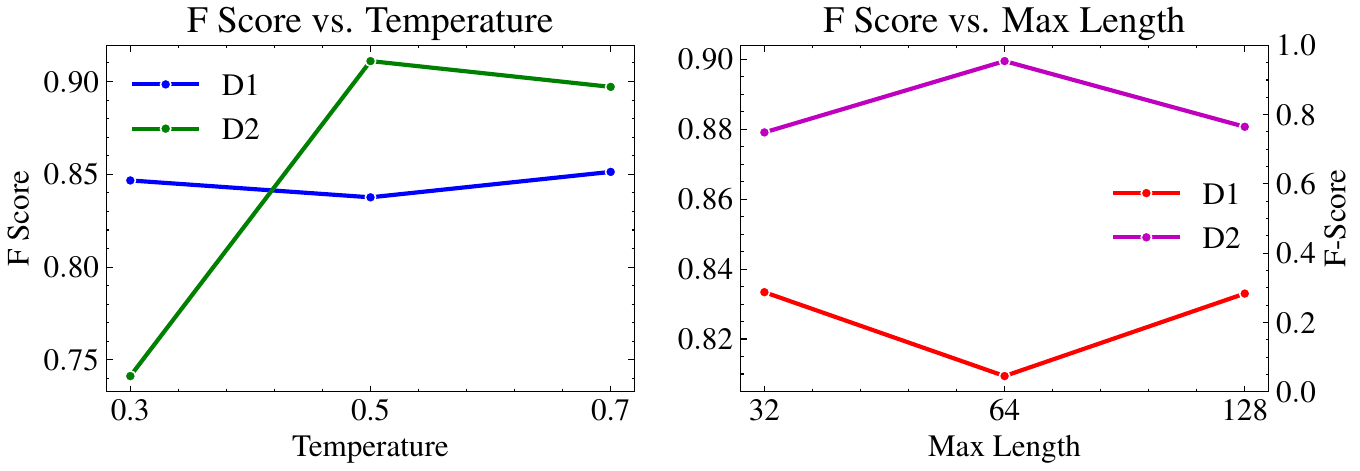}\label{fig:par_ml}} 
 \caption{Parametric analysis}
 \label{fig:parametricanalysis}
 \vspace{-15pt}
\end{figure}

We examine the influence of five parameters on \textsc{DepressionX} - learning rate, number of heads for multi-head attentions, number of heads in GAT, temperature value for contrastive loss, and maximum number of words - shown in Figures \ref{fig:par_an} and \ref{fig:par_ml}.

Figure \ref{fig:par_an} demonstrates the effects of learning rate and number of heads used in multihead attention and GAT, on the performance \textsc{DepressionX}. \textbf{D1} is more responsive to changes in learning rate as compared to \textbf{D2}. This implies that the ideal learning rate could differ according to the specific properties of the dataset, such as the distribution of labels. We observe that adjustments to the number of heads in the GAT architecture have a lesser effect compared to the multi-head attention model. Increasing the number of heads generally results in enhanced performance up to a certain threshold. Beyond this point, further increases may lead to diminishing returns or even a decline in performance, as indicated by fluctuations in the $\text{F}_1$ score.

Figure \ref{fig:par_ml} presents the parametric analysis of the temperature values for contrastive loss and the maximum number of words within sentences considered for embedding. The $\text{F}_1$ score is clearly influenced by the temperature value, with the highest $\text{F}_1$ score obtained at a temperature of 0.7 for \textbf{D1} and 0.5 for \textbf{D2}. This suggests that the optimal temperature value may vary based on the properties of the dataset. Regarding the maximum length of words, we observe a similar trend, where increasing the maximum length generally leads to improved performance up to a certain point before performance degrades.

\section{Conclusion} \label{sec:conclusion}

In this paper, we introduced \textsc{DepressionX}, a novel knowledge-infused explainable model for predicting the severity of depression from textual expressions in social media posts. \textsc{DepressionX} comprises two main modules: post analysis and domain-specific KG analysis. The first module uses word-level, sentence-level, and post-level embeddings to generate a rich representation of social media posts with contextual embeddings. The second module incorporates knowledge from Wikipedia and the BDI-II to construct a depression KG, which is then processed with GIN and GAT architectures to enhance our model's knowledge awareness. The final post representation is created by infusing the textual and KG representations, which is then used to classify the severity of depression. \textsc{DepressionX} achieved $\text{F}_1$ scores of 82.5\% and 90.9\% on imbalanced and balanced datasets, respectively, surpassing the compared models by more than 7\% on both datasets. Our model is expected to have a significant impact in the field of AI applied to mental health.
Additionally, there are a few limitations that must be taken into account. Our methodology exclusively concentrates on the textual content of social media posts, disregarding other potential multimodal features, including behavioural information. Furthermore, the model's insights could be further enhanced by considering the social interconnections among users.

\section*{Ethics Approval}
We maintained strict ethical standards throughout our study. Our data analysis is approved by \textbf{Physical Sciences Ethics Committee of the University of York} under the application reference \textbf{Ibrahimov20230330}. No personally identifiable information were accessible in this research.

\bibliography{aaai25}

\end{document}